\documentclass[letterpaper]{article} 
\usepackage{aaai25}  
\usepackage{times}  
\usepackage{helvet}  
\usepackage{courier}  
\usepackage[hyphens]{url}  
\usepackage{graphicx} 
\usepackage{natbib}  
\usepackage{caption} 
\usepackage{algorithm}
\usepackage{algorithmic}

\usepackage{newfloat}
\usepackage{listings}
\DeclareCaptionStyle{ruled}{labelfont=normalfont,labelsep=colon,strut=off} 
\floatstyle{ruled}
\newfloat{listing}{tb}{lst}{}
\floatname{listing}{Listing}
\usepackage{amsmath}
\usepackage{makecell}
\usepackage{booktabs}

\title{Are AI Detectors Good Enough? A Survey on Quality of Datasets with Machine-Generated Texts}
\author {
    German Gritsai\equalcontrib\textsuperscript{\rm 1},
    Anastasia Voznyuk\equalcontrib\textsuperscript{\rm 2},
    Andrey Grabovoy\textsuperscript{\rm 1},
     Yury Chekhovich \textsuperscript{\rm 1}
}
\affiliations {
    \textsuperscript{\rm 1}Advacheck OÜ, Tallinn, Estonia\\
    \textsuperscript{\rm 2}Moscow Institute of Physics and Technology, Moscow, Russia\\
    gritsai@advacheck.com,
    vozniuk.ae@phystech.edu
}

\begin{document}

\maketitle

\begin{abstract}
The rapid development of autoregressive Large Language Models (LLMs) has significantly improved the quality of generated texts, necessitating reliable machine-generated text detectors. A huge number of detectors and collections with AI fragments have emerged, and several detection methods even showed recognition quality up to 99.9\% according to the target metrics in such collections. However, the quality of such detectors tends to drop dramatically in the wild, posing a question: Are detectors actually highly trustworthy or do their high benchmark scores come from the poor quality of evaluation datasets? In this paper, we emphasise the need for robust and qualitative methods for evaluating generated data to be secure against bias and low generalising ability of future model. We present a systematic review of datasets from competitions dedicated to AI-generated content detection and propose methods for evaluating the quality of datasets containing AI-generated fragments. In addition, we discuss the possibility of using high-quality generated data to achieve two goals: improving the training of detection models and improving the training datasets themselves. Our contribution aims to facilitate a better understanding of the dynamics between human and machine text, which will ultimately support the integrity of information in an increasingly automated world.
\end{abstract}

%

\section{Introduction}
The quality of large language models (LLMs) has grown tremendously in the last five years, making their output almost indistinguishable from human-written texts ~\cite{llmsprogress}.
This expanded the application fields of these models, as many routine tasks can be entrusted to them nowadays. However, they can be used for creating texts that are intended to be written and fact-checked by humans. An example of such misuse is the generation of fake news~\cite{zellers2020defending, Synthetic_Lies}, which can mislead readers of such generated content. Teachers raise another concern, as many students complete assignments with LLMs~\cite{outfox, ma2023ai}, undervaluing the purpose of the educational process. Machine-generated fragments also appear in academic articles more often with the growth of chatbots and reach several tens of percent ~\cite{liang2024mapping, gritsay2023artificially}. More than 60,000 scientific papers in the last year alone contained evidence of the use of machine generation~\cite{gray2024chatgpt}. All of that proves that it is crucial to develop systems able to counter the misuse of artificial data and signal to the reader that the content they read is generated.

Another concern is the Web, overflowing with machine-generated content, often of poor quality. Such texts contribute bias to publicly available texts on the Internet, through false facts, hallucinations and spelling errors. Given the current agenda of using texts from the Internet to train new language models, all this bias will be inadvertently added to the model. Moreover, ~\cite{villalobos2022will} revealed that the human-written data will run out by 2028. That means that the training sets for language models in the future will include a large amount of generated content. Such \textit{self-consuming} will result in the substantial degradation of the model's abilities~\cite{self-consuming}. Furthermore, the trend is evolving in such a way that human-written texts on almost any topic will be much harder to retrieve. While for texts dated even 5 years ago we are confident as the usage of generation was extremely rare, we cannot state the same for more recent texts.

Therefore, detectors capable of distinguishing human-written texts from AI-generated texts and whose detection quality can be guaranteed, are necessary for many fields. We believe that one of the key factors for building reliable detectors is the high-quality artificial text collections that can be used for training and evaluation. In this paper, we would like to estimate the quality of the available generated texts from competitions and research papers. Sometimes we see that some methods from participants of the competitions reach almost perfect (up to 99.9\%) metric score, meanwhile in the wild we observe a noticeable decline in performance. Such results look confusing, because the models become more and more advanced, seemingly making the detecting task more challenging, meanwhile participants of competitions still reach almost perfect scores, bringing up the question about quality of generated data in the provided datasets. Are the devised methods really good or is the data easy enough for detectors to solve the seemingly hard detection task?

Our contributions are as following
\begin{enumerate}
    \item We systemize information about existing datasets from the research papers and competitions, dedicated to the detection of AI-generated content task.
    \item We suggest methods that may be helpful for evaluating the quality of the generated data and the datasets aimed to use for binary classification between human and machine texts.
\end{enumerate}

\section{Related Work}

\subsection{The Task of AI-generated text detection}
\label{section:task_methods}

The task of AI-generated text detection task is generally stated as a text classification task, which means that the input is a text sequence and the output is a discrete, usually binary, class prediction. When the task is binary, the common labels are “AI” or “human”, whereas multiclass classification focusses on distinguishing several language models. The last task is usually called authorship attribution. Finally, more complex task suggests to determine the borders betweens fragments from different authors, for example between human author and some LLM author.

The first approaches to tackle the classification problem were to utilise some linguistic, stylometric, and statistical features for classifiers~\cite{jawahar-etal-2020-automatic, Frhling2021FeaturebasedDO}. However, while these methods performed well for the texts from the first language models, nowadays models are advanced enough to output texts that are almost indistinguishable from human-written ones, therefore these methods are currently not reliable enough. The next category are zero-shot methods that employ metrics, such as perplexity or its modifications~\cite{hans2024spotting}, which can be helpful as an inference method when training is not available. Another approach that do not require training is perturbing texts, which can also provide valuable information. For example, one can compare log-probabilities between between original and perturbed texts, as described in DetectGPT~\cite{mitchell2023detectgpt} method. Finally, methods based on fine-tuning encoder-based models, such as DeBERTa~\cite{he2021deberta}, are currently considered the state-of-the-art approach for the detection task~\cite{uchendu2021turingbench, macko-etal-2023-multitude}.

\subsection{Evaluating Generated Text}
As for evaluating the quality of the generated data itself, it has become more common to evaluate it with the help of LLMs~\cite{INSTRUCTSCORE}. This approach does not require any human reference, unlike ROUGE~\cite{rouge}. However, the output of model-evaluator needs to be unified, is not always interpretable, and model-evaluator scores can be skewed. Alternative approach is suggested by ~\cite{zhu-bhat-2020-gruen}, where the text is evaluated based on several linguistic criteria, such as grammar or coherence.

\subsection{Datasets with Artificial Content}
There are a number of surveys of machine-generated content detection with an overview of the datasets~\cite{jawahar-etal-2020-automatic, wu2023survey}, however, few works focus on the quality of data in the available datasets, despite it being an important aspect of the task. Building AI-generated content detectors requires high-quality labelled data that involve substantial financial, computational, and human resources. The human evaluators should check that the dataset does not contain corrupted generations, that the texts are coherent and grammatically correct. We will describe the datasets we used in our analysis and experiments in Section~\ref{section:data}.

\subsection{Shared Tasks on AI-Generated content Detection}

Shared tasks advance research on detecting AI-generated content forward by offering new variations on tasks and providing data for evaluation, encouraging participants to come up with novel ideas for detectors robust to the change of language, domain, or generating model. Participants explore approaches ranging from transfer learning on complex text features to utilising and fine-tuning LLMs for these tasks~\citeyearpar{ruatd, dg22, owniber, boeva, dp24, acl, panwin, iilmas}. These efforts have highlighted challenges such as handling multilingual data and adapting to rapidly evolving generative models. Some participants also provide some analysis of the given data or even discuss some flaws with the generated texts~\cite{voznyuk-konovalov-2024-deeppavlov}.

\section{Datasets}
\label{section:data}

\renewcommand{\arraystretch}{1.1} 
\begin{table*}[t]
\centering
\scriptsize
\resizebox{\textwidth}{!}{%
\begin{tabular}{@{}p{1.8cm}|p{1.6cm}<{\centering} p{1.2cm}<{\centering} p{1.7cm}<{\centering} p{1.9cm}<{\centering} p{1.9cm}<{\centering}@{}}
\toprule
\textbf{Dataset}  & \textbf{Language} & \textbf{Num. of Texts, $10^3$} & \textbf{Num. of Texts, G / H, $10^3$} & \textbf{Average Length, G \textbf{/} H} & \textbf{Median Length, G \textbf{/} H} \\ 
\midrule
GPT2 & en & 1250 & 1000 \textbf{/} 250 & 2941 \textbf{/} 2616 & 3245 \textbf{/} 2459 \\
\hline
TweepFake  & en & 20.7 & 10.4 / 10.4 & 104 / 118 & 89 / 94 \\
\hline
HC3 & en, zh & 85.4 & 26.9 \textbf{/} 58.5 & 1011 \textbf{/} 681 & 1012 \textbf{/} 422 \\
\hline
GhostBuster  & en & 21 & 18 \textbf{/} 3 & 3345 \textbf{/} 3391 & 3440 \textbf{/} 2911.5 \\
\hline
MGTBench  & en & 23.7 & 20.7 \textbf{/} 3 & 1596 \textbf{/} 3391 & 1226 \textbf{/} 2911.5 \\
\hline
MAGE &  en & 436 & 152.3 \textbf{/} 284.2 & 1139 \textbf{/} 1282 & 706 \textbf{/} 666 \\
\hline
M4 &  en, zh, ru, bg, ur, id  & 89.5 & 44.7 \textbf{/} 44.7 & 1588 \textbf{/} 3162 & 1454 \textbf{/} 1697 \\
\hline
OutFox  & en & 57.6 & 43.2 / 14.4 & 2686 / 2238 & 2311 / 1992 \\
\midrule
DAGPap22  & en & 5.3 & 3.6 \textbf{/} 1.6 & 799 \textbf{/} 1180 & 680 \textbf{/} 1126.5 \\
\hline
RuATD  & ru & 129 & 64.5 \textbf{/} 64.5 & 237 \textbf{/} 221 & 99 \textbf{/} 95\\
\hline
AuTex  & en, es & 65.9 & 33.1 \textbf{/} 32.8 & 315 \textbf{/} 297 & 386 \textbf{/} 351 \\
\hline
IberAuTex  & es, en, ca, gl, eu, pt & 98 & 52.5 \textbf{/} 45.4 & 1037 \textbf{/} 1058 & 981 \textbf{/} 1018 \\
\hline
PAN24 & en & 15.2 & 14.1 \textbf{/} 1.1 & 2641 \textbf{/} 3007 & 2731 \textbf{/} 2868 \\
\hline
SemEval24 Mono & en & 34.2 & 18 \textbf{/} 16.2 & 2465 \textbf{/} 2358 & 2570 \textbf{/} 2083.5 \\
\hline
SemEval24 Multi & en, ar, de, it & 42.3 & 22.1 \textbf{/} 20.2 & 2218 \textbf{/} 2257 & 2270 \textbf{/} 2032 \\
\hline
MGT-1 Mono & en & 610.7 & 381.8 \textbf{/} 228.9 & 1448 \textbf{/} 1541 & 1208\textbf{/} 1080 \\
\hline
MGT-1 Multi & en, zh, it, ar, de, ru, bg, ur, id & 674 & 416.1 \textbf{/} 257.9 & 1423 \textbf{/} 1445 & 1195 \textbf{/} 1032 \\
\bottomrule
\end{tabular}
}
\caption{Statistics of the texts in the datasets from the shared tasks and research papers.}
\label{table:info_datasets}
\end{table*}

\subsection{Datasets From Shared Tasks}

 The most common tasks in shared tasks are binary classification and authorship attribution, with binary classification being the prevalent task, therefore, in this work, we focused only on it. 
All chosen shared tasks contain texts in English, unless stated otherwise. Here we give a brief overview of each task, as well as some quantitative statistics of the texts in Table~\ref{table:info_datasets}, whereas a more detailed description, such as models used for generation or domains of the presented texts, can be found in Appendix \ref{appendix:data}.

\begin{itemize}
\item \textbf{DAGPap 2022}~\cite{kashnitsky-dagpap2022} introduced a dataset of human- and machine-written scientific excerpts collected by Elsevier.
\item \textbf{RuATD 2022}~\cite{ruatd} focused on human- and machine-written documents in Russian, covering a wide range of themes.
\item \textbf{AuTexTification 2023}~\cite{AuTexTification2023} provided texts in English and Spanish, covering five distinct domains.
\item \textbf{IberAuTexTiification 2024}~\cite{iberautextification} expanded on the previous competition with a multilingual (six Iberian languages), multi-domain, and multi-model focus.
\item \textbf{Voight-Kampff Generative AI Authorship Verification 2024}~\cite{pan2024}, hereafter referred to as PAN 2024, tasked participants with identifying the human-authored text from two samples -- one human-written and one machine-generated.
\item \textbf{SemEval 2024 Task 8}~\cite{semeval2024task8} addressed domain, generator, and language shifts in generated texts. Training data included multiple languages such as Chinese, Urdu, and Russian, but the test set was limited to English, Italian, German, and Arabic.
\item \textbf{MGT Detection Task 1 (COLING 2025)}~\cite{wang2025genai} was built on SemEval 2024 Task 8 by incorporating data generated by novel LLMs and expanding the multilingual coverage of the train and test sets.

\end{itemize}

\begin{figure*}[t]
\centering
\includegraphics[width=0.90\textwidth]{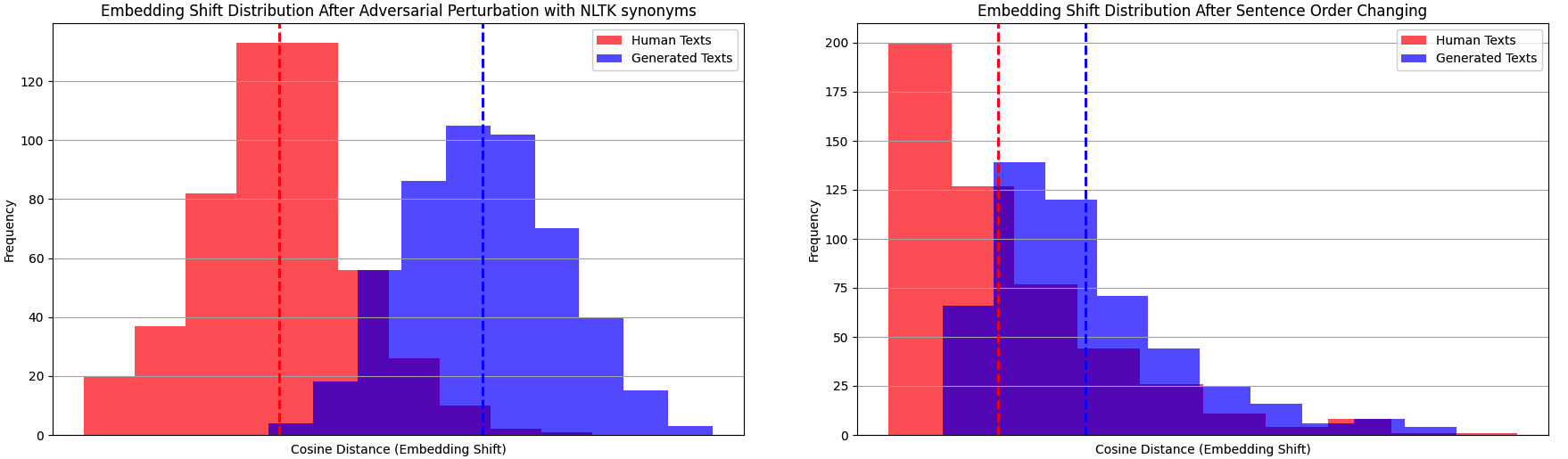}
\caption{Comparison of embedding shifts after two types of modifications for the HC3 dataset.}
\label{fig::pertrub}
\end{figure*}

\begin{figure*}[t]
\centering
\includegraphics[width=1.00\textwidth]{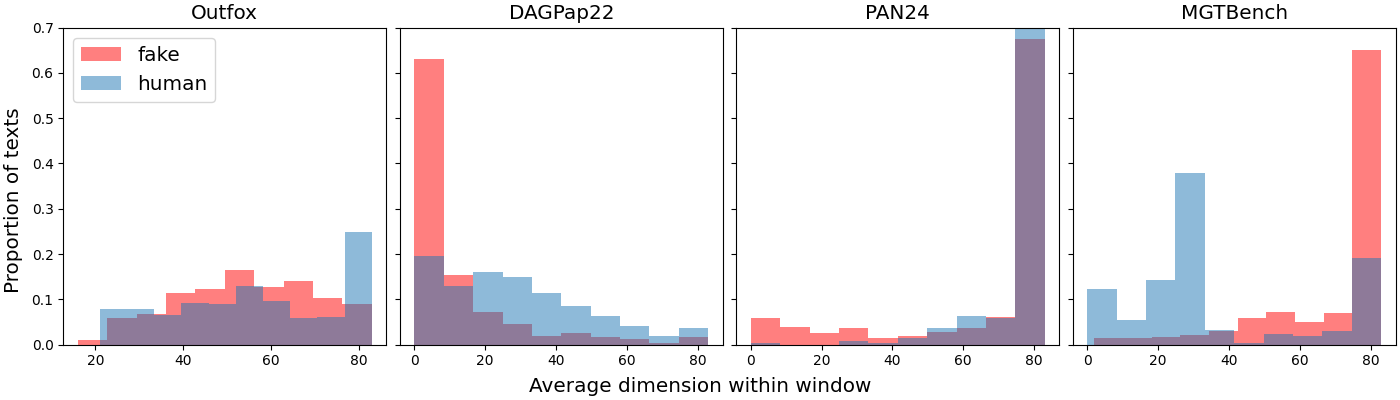}
\caption{Topological Time Series for different datasets. The results for the remaining datasets selected in this paper can be found in Figure~\ref{fig:tts}.} 
\label{tts}
\end{figure*}

\subsection{Datasets from Research Papers}
The number of collections with generated content has started to grow with an increasing number of available generators. Quite often, researchers, together with a new approach for AI content detection, publish a parallel dataset on which they have validated their method. In this paper, our aim was to pick collections with human- and machine-generated excerpts that are the most common and cited in other researchers' publications.
Similarly to previous subsection, here we give a brief overview of each chosen datasets, describe some statistics about the texts in Table~\ref{table:info_datasets}, and add a more detailed description in Appendix~\ref{appendix:data}.

\begin{itemize}
\item \textbf{GPT2 Output Dataset}\footnote{https://github.com/openai/gpt-2-output-dataset} consists of text outputs generated by GPT-2 models of different sizes across various prompts.
 
\item \textbf{HC3}~(Human Chatbot Conversations Corpus)~\cite{Su2023HC3PA} features conversations between humans and chatbots, primarily used for research on chatbot responses and human-AI interaction analysis. This dataset is available for both English and Chinese, but we have focused only on the former.

\item \textbf{GhostBuster}~\cite{Ghostbuster} aimed at detecting AI-generated content by comparing it to human-written text, often used in the context of identifying machine-generated misinformation or spam.

\item \textbf{MGTBench}~(Machine Generated Text Benchmark)~\cite{MGTBench} is a benchmark dataset designed to evaluate the quality of machine-generated text across various tasks, including fluency, coherence, and creativity.
 
\item \textbf{MAGE}~(Model Augmented Generative Evaluation)~\cite{li2024mage} evaluates the performance of generative models by comparing outputs with human annotations, aiding in the development of more accurate generative AI models.

\item \textbf{M4}~(Multilingual, Multimodal, Multitask, Massive Dataset)~\cite{wang2023m4} is a large-scale dataset designed for training models that can handle multiple languages, tasks, and modalities, making it useful for developing versatile AI systems. Although it is multilingual, we sampled only English texts.

\item \textbf{TweepFake}~\cite{Fagni_2021} contains real tweets written by humans and synthetic tweets, generated by various AI models, from bots, imitating human users.

\item \textbf{Outfox}~\cite{outfox} contains triplets of essay problem statements, human-written essays, and LLM-generated essays. The students who wrote the essays range from 6th to 12th grade in the USA.
\end{itemize}

\section{Approach}
We decided to evaluate all datasets with common setups to see how good standard approaches perform on them. We did not have the goal to obtain the highest score, but rather to compare the performance of the same method on different datasets.

\subsection{Baselines}
In Section \ref{section:task_methods} we described three main categories of methods for tackling the detection task. We chose a method from each category, that served as a baseline to obtain first-hand understanding of each dataset. For the perturbation-based methods we used \textbf{DetectGPT} framework with GPT-2 \cite{gpt2} as the base model and T5-Large \cite{t5} as perturbations generator. However, due to intensive computational costs of DetectGPT, we utilised Fast-DetectGPT~\cite{fast-detectgpt} that substitutes DetectGPT's perturbation step with a more efficient sampling step. For the zero-shot methods we used \textbf{Binoculars}~\cite{hans2024spotting} with improved perplexity score. These two baselines need no fine-tuning, which is an important aspect for detection task, as it is infeasible to train the detector for every domain and generator. Lastly, as encoder-based method we used 
\textbf{mDeBERTa}~\cite{he2021deberta}, which is the current state-of-the-art model for multilingual machine-generated text detection~\cite{macko-etal-2023-multitude}. By taking these three detectors, we covered all main categories of detectors.

\subsection{Topological Statistics}
It was shown in \cite{Tulchinskii_phd} that if we take the inner dimensionality of the manifold on the set of embeddings, we could separate human-written texts from machine-generated ones. The authors used persistence homology dimension (PHD) and showed that statistically human-generated texts have higher PHD than machine-generated texts, therefore introducing a novel detector. We calculated PHD on each set of texts. Additionally, in ~\cite{kushnareva2024boundary} it was suggested to calculate PHD within sliding window. These intrinsic dimensions of the text within sliding window can be used as a feature for detectors.  The authors show that the metric is robust to the change of domain and generators. To be able to compare datasets between each other, we came up with a symmetrical score, utilising KL-divergence. Let $h_d$, $m_d$ be distributions of intrinsic dimensions for two types of texts from the same dataset, of human and machine origin, then our $\text{KL}_{\text{TTS}}$ is following:
$$\text{KL}_{\text{TTS}} (h_d, m_d) = | D_{\text{KL}}(h_d || m_d) - D_{\text{KL}}(m_d || h_d) |$$

The lower this score, the closer $h_d$ and $m_d$ are, which means almost indistinguishable texts and vice versa.

\subsection{Perturbations and Shuffling}
Based on the results of the text modification studies~\cite{sadasivan2024can, mitchell2023detectgpt}, which show how small perturbations affect machine reading comprehension systems, we decided to consider this way of possibly assessing the quality of a dataset. The key idea here is that AI models are sensitive to such adversarial changes, unlike humans. We considered two modification ideas: Adversarial Token Perturbation and Sentence Shuffling.

\textbf{Adversarial Token Perturbation.} In this approach we divide the text into tokens and randomly replace the token with a synonym from the WordNet collection~\cite{wordnet} with a probability of 70\%. We apply such a technique to each represented class. Using an encoder model, we obtain embeddings for each of the texts in the current dataset. Finally, we measure the average embedding shifts for the classes of human and generated texts. We obtain the embedding shifts using the cosine distance between the embeddings of the original texts and the modified ones. As a result, after modifications we obtain $\Delta_{\text{shift}}$ --- the log difference of the average embedding shifts.

\[
\Delta_{\text{shift}} = \log \frac{{\frac{1}{n} \sum_{i=1}^{n} \text{cos}_d(h_{h_i}^o, h_{h_i}^p)}}{{\frac{1}{m}\sum_{j=1}^{m} \text{cos}_d(h_{m_j}^o, h_{m_j}^p)}},
\]


where $n$ and $m$ are number of samples in the human and generated parts of the dataset respectively, $h_{h_i}^o$ -- embedding of the $i$-th fragment of human part of data, $h_{h_i}^p$ -- the same embedding after perturbation. Similarly, $h_{m_i}^o$ and $h_{m_i}^o$ are embeddings for machine-generated texts. Finally, $\text{cos}_d$ is a function that measures the cosine distance between two vectors.

\textbf{Sentence Shuffling.}
In this approach, we randomly swap sentences, thereby affecting the cohesion of the text. We try to find out the effect of artificial origin on the difference between the distributions after permutations. By dividing a fragment into sentences and randomly reversing the order of 70\% of the selected sentences, we apply this technique to each represented class. Then, using the text encoding model, we obtain embeddings for each of the texts from the current dataset. Finally, we measure embedding shifts for the class of human and generated texts, and after that we convert the shifts into probability-like distributions. This allows us to obtain at the end $\text{KL}_{\text{shuffle}}(H,M)$ --- the KL-divergence between the shifts of human and generated texts.

\[
\text{KL}_{\text{shuffle}}(H,M)= \sum_{i} H(i) \log \frac{H(i)}{M(i)},
\]
\[
H(i) = \frac{\text{cos}_d(h_{h_i}^o, h_{h_i}^p) + \epsilon}{\sum_j \left(\text{cos}_d(h_{h_j}^o, h_{h_j}^p) + \epsilon\right)}. 
\]
$M(i)$ has the same structure as $H(i)$, except that instead of human class texts the generated class texts are used, $\epsilon$ is a small constant added to avoid division by zero.

\section{Experiments}
\label{section:experiments}

\renewcommand{\arraystretch}{1.2} 
\begin{table}[t]
    \centering
    \begin{tabular}{p{1.8cm}|p{1.5cm}<{\centering}|p{1.5cm}<{\centering}|p{1.4cm}<{\centering}}
    \toprule    
        \textbf{Dataset} & \textbf{DeBERTa} & \textbf{Binoculars} & \textbf{DetectGPT} \\
        \midrule
        GPT-2 &  0.972 & 0.495 & 0.412\\ 
        HC3 & 0.998 &  0.931 & 0.972 \\
        GhostBuster & 0.910 &   0.683  &  0.711 \\
        MGTBench & 0.961 & 0.364 & 0.447 \\
        MAGE &  0.835 &  0.632 &  0.654 \\
        M4 & 0.987 &  0.871 &  0.881   \\
        OutFox & 0.901 & 0.692 & 0.707 \\
        TweepFake & 0.941 & 0.845 & 0.864 \\ 
        \midrule
        \makecell[l]{SemEval24 \\ Mono} & 0.991 & 0.913 & 0.924 \\
        \makecell[l]{SemEval24 \\ Multi} & 0.994 & -- & --\\
         RuATD & 0.765 & -- & -- \\
         DAGPap22 &  0.968 & 0.333 & 0.562 \\
         PAN24  & 0.826 &  0.411 & 0.890 \\
        \makecell[l]{AuTex23en} & 0.941 & 0.783 & 0.911\\
        \makecell[l]{AuTex23es} & 0.933 & -- & --\\
        IberAuTex & 0.964 & -- & --\\
        \makecell[l]{MGT-1\\Mono} & 0.904 & 0.665 & 0.683 \\
        \makecell[l]{MGT-1 \\ Multi} & 0.934 & -- & --\\
    \bottomrule    
    \end{tabular}
    \caption{Classification results with different detectors estimated using $F_1$-score. Binoculars and DetectGPT work only with English texts, thus we could not apply them to datasets with non-English texts.}
    \label{tab:classifier}
\end{table}

\begin{table*}[t]
    \centering
    \begin{tabular}{l|c|c|c|c|c}
    \toprule
        Dataset & $\text{KL}_{\text{TTS}}$ $\downarrow$ & $\text{PHD}_{\text{human}}$ & $\text{PHD}_{\text{machine}}$ &  $\Delta_{\text{shift}}$ $\downarrow$ &  $\text{KL}_\text{shuffle}$ $\downarrow$  \\
        \midrule
        GPT-2 & \textbf{0.014} & 9.23 $\pm$ 1.98 & 10.27 $\pm$ 1.84 &  0.084 & 1.255  \\
        HC3 & 0.053  & 8.76 $\pm$ 1.83 & 7.38 $\pm$ 1.05 & 0.264 &   1.167  \\
        GhostBuster & 0.053 & 9.84 $\pm$ 1.18 & 9.76 $\pm$ 1.15 & \textbf{0.024} &  \textbf{0.359}  \\
        MGTBench & 0.043  & 8.77 $\pm$ 1.31 & 9.97 $\pm$ 1.02 &  \textbf{0.031} &   \textbf{0.421} \\
        MAGE & \textbf{0.011}  & 9.8 $\pm$ 2.14 & 9.38 $\pm$ 3.04 & 0.094  & \textbf{0.310}  \\
        M4 & 0.036 & 7.26 $\pm$ 1.99 & 8.59 $\pm$ 1.4 & 0.107 &  \textbf{0.483}  \\
        OutFox & 0.025  & 8.96 $\pm$ 1.21 & 11.48 $\pm$ 1.13  & 0.095 & \textbf{0.237} \\
        TweepFake & \textbf{-} & 9.02 $\pm$ 3.19 & 8.12 $\pm$ 4.02 & 0.116 & 1.001 \\
        \midrule
        SemEval24 Mono & \textbf{0.012} & 9.11 $\pm$ 1.19 & 9.41 $\pm$ 1.2 &    0.191 &  2.576  \\
        SemEval24 Multi & \textbf{0.001}  & 9.65 $\pm$ 1.81 & 9.42 $\pm$ 1.44 &  0.059 &  2.046  \\
         RuATD & \underline{0.007} & 7.33 $\pm$ 1.4 & 7.46 $\pm$ 1.41 &  0.315 &  14.028  \\
         DAGPap22 & 0.083 & 8.35 $\pm$ 1.33 & 7.48 $\pm$ 2.01 &  \textbf{0.039} &   \textbf{0.472}  \\
          PAN24 & 0.053 & 9.4 $\pm$ 1.05 & 8.52 $\pm$ 1.59 &  \textbf{0.050} &  \textbf{0.331}  \\
          AuTex23 Eng & \underline{0.021} & 8.07 $\pm$ 2.26 & 8.1 $\pm$ 2.68 &  0.110 &  4.331 \\
          AuTex23 Esp &  \underline{0.001}& 9.16 $\pm$ 3.49 & 9.25 $\pm$ 3.26  &   0.105  &   1.306 \\
          IberAuTex & \textbf{0.012} & 9.33 $\pm$ 2.45 & 8.47 $\pm$ 2.73 &  0.223 & 5.516 \\
          MGT-1 Mono & \textbf{0.019} & 9.19 $\pm$ 1.75 & 8.96 $\pm$ 2.24 & \textbf{0.031} & 0.587\\
          MGT-1 Multi & \textbf{0.006} & 
8.76 $\pm$ 1.85 & 8.6 $\pm$ 2.29 &  \textbf{0.027} & 0.522\\
    \bottomrule    
    \end{tabular}
    \caption{Calculated statistics on texts from chosen datasets. Some values for $\text{KL}_{\text{TTS}}$ are underlined, because texts are too short, see Section~\ref{sec:discussion} and TTS for almost all texts in TweepFake is equal to 0.}
    \label{tab:results}
\end{table*}

From each dataset, we sampled 1000 documents from the test set, balanced between two classes. Regarding baselines, we fine-tuned \texttt{mdeberta-v3-base} for each dataset and evaluated the model. Additional information about hyperparameters during training can be found in the Appendix \ref{appendix:hyperparameters}. To evaluate the quality of Binoculars and Fast-DetectGPT, we utilised \texttt{falcon-rw-1b}~\cite{almazrouei2023falconseriesopenlanguage} and \texttt{gpt-neo-2.7B}~\cite{gpt-neo} respectively. It is worth noting that with the last two methods we were only able to measure quality for samples in English. 

Our objective was to show that datasets of lower quality have shifts that will be easily recognised by the models "from the first step", hence we have not performed any hyperparameter tuning, only one iteration of fine-tuning and testing of the underlying models. In the experiment with topological features we used \texttt{roberta-base}, just as the authors of the original paper. In the experiment with perturbations and shuffling, the \texttt{multilingual-e5-large} encoder was used to build embeddings of texts, which shows high metrics on encoding high-resource languages~\cite{e5}.

\section{Results}
\label{section:results}

\begin{figure*}[ht!]
\centering
\includegraphics[width=\textwidth]{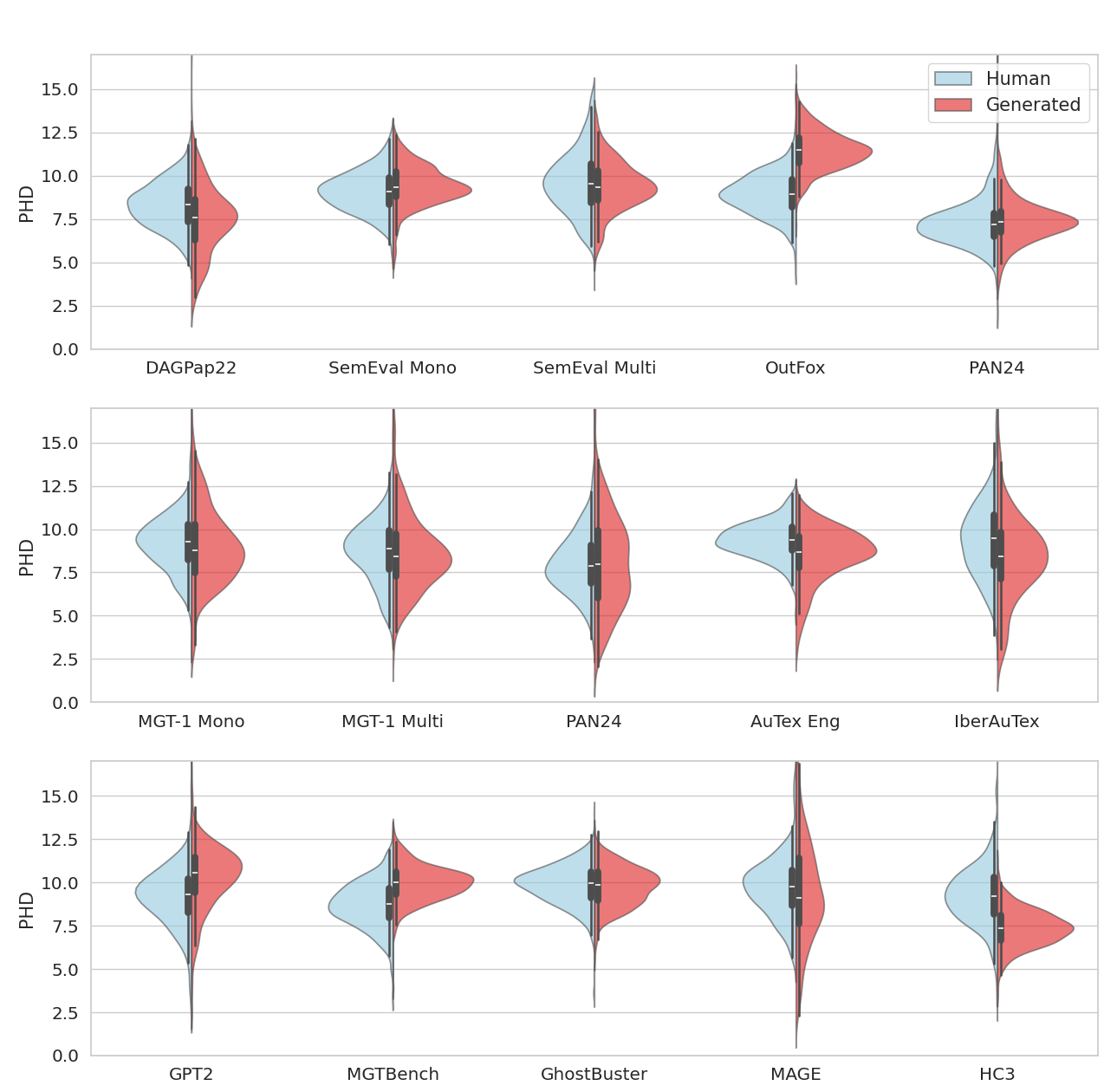}
\caption{PHD values on all datasets, except TweepFake and AuTex23 Spanish, texts from which were too short for proper calculation of PHD.}
\label{fig:phd_values}
\end{figure*}

The results of the comparison of the designed features in the selected datasets are presented in Table~\ref{tab:results}. 
Regarding the PHD and TTS score, in previous works it was shown that texts from language models have smaller PHD values than human-written ones; however, this result was obtained for GPT-2, GPT-3.5 and OPT models, and this trend could change for more recent language models that generate more human-resembling texts. If texts of different origin have high $\text{KL}_{\text{TTS}}$, it means that it is easier for a detector to separate such texts. $\text{KL}_{\text{TTS}}$ is also constrained for shorter texts, see Section~\ref{sec:discussion}. As for PHD, we hypothesise that generated texts of good quality should have PHD similar to human-written ones. Additionally, we compare the distributions of PHD for all datasets on Figure~\ref{fig:phd_values}. Again, the distributions for texts of both origin should be similar, which mainly holds for texts from SemEval, PAN24 and MGT-1.

In the next columns, we list the statistics observed on modified texts, and for both of these the lower the better, as this reflects the similar degree of resilience of the generated and human texts to adversarial attacks. Qualitatively generated data with no bias should take values close to human. 

Finally, in Table~\ref{tab:classifier} we show the results of applying modern detectors to the chosen test datasets. For instance, on the datasets with low values in Table~\ref{tab:results}, a quality close to 1 can be achieved, indicating the clear presence of detector bias towards them, or a structural feature that is too obvious for the detection model. It is not possible to judge the quality of the data only by achieving $F_1$ values close to 1, but by combining the values of the two tables, we can estimate which set has better quality data and which has lower quality data. 

\section{Discussion}
\label{sec:discussion}

Regarding $\text{KL}_{\text{TTS}}$, on Fig.~\ref{tts} we show 4 datasets with the high value of it. While GhostBuster and PAN24 received such a high score due to the discrepancy on texts with higher dimensions, MGTBench and DAGPap22 did it due to the difference in distributions themselves. Note also that $\text{KL}_{\text{TTS}}$ may not perform well with very short texts, since the internal method of computing PHD requires sufficiently long texts for stable computation. Therefore, we discard $\text{KL}_{\text{TTS}}$ on RuATD and AuTex23-es and Tweepfake, as they do not fit the criteria; see Table~\ref{table:info_datasets}. In addition, it has already been shown that the texts must be of sufficient length~\cite{need_more_tokens} to build reliable detectors.

Analysing the values from Table~\ref{tab:results}, we can trace the presence of sufficiently high quality data in the selected datasets. The developed attributes in aggregate are able to reflect the quality of the generated dataset from different perspectives and angles. We propose to utilise these attributes in combination with other statistical tools to evaluate data quality, for example, Zipf's law~\cite{zipf}.

Presented statistics can be utilised to estimate the quality of collections and to improve them. In addition, datasets that collect machine-generated content may also provide utility for the two more general purposes. First, high-quality generated data can be utilised to evaluate the quality of the causal model during training, as one of the training objectives to improve model answers and make it more human-like. Secondly, good detectors can help to clean training sets, as large proportion of low-quality generated texts in those sets can result in emerging biases towards incorrect structure and rubbish fragments in the output of the model in the future.

The question of whether poor performance by detectors implies poor dataset quality yields an ambiguous answer. For instance, in \cite{hans2024spotting}, the Binoculars method achieves an F$_1$-score close to 1.0, while our experiments produced a wide range of scores: from 0.33 on DAGPap22 to 0.93 on HC3. For HC3, all three detectors performed similarly, suggesting that the HC3 texts are relatively easy to detect. However, this consistency does not extend to DAGPap22. For instance, the DeBERTa-based detector achieved an F$_1$-score of 0.96, while DetectGPT scored only 0.562. This pattern, where the DeBERTa-based detector achieves notably higher scores than the other two methods, was observed across a significant portion of the analysed datasets. We attribute this strong performance to the fine-tuning of the DeBERTa-based detector.

Conversely, the low scores for Binoculars merit further scrutiny. Even when focusing on domains specifically tested by its authors, such as PAN24 (News) and Outfox (Student Essays), the scores fall well below the near-perfect results reported in \cite{hans2024spotting}. This discrepancy suggests that the Binoculars detector may not be representative. Similarly, in our experiments, DetectGPT’s scores are comparable to Binoculars’ scores, potentially indicating similar underlying issues with the robustness of these detectors.

\section{Conclusion} 
In the current research, we discussed the problem of quality of datasets with AI-generated texts used for testing corresponding detectors. This problem is relevant, as the quality of test data directly influences the quality of widely used detectors. We conducted a review of datasets from competitions and scientific publications on datasets aimed at the detection of AI-generated content and proposed methods to evaluate the quality of datasets containing AI excerpts based on different structural features. We evaluated topological features, robustness to adversarial attacks, and performance of the widely used detectors on these datasets. We concluded that all analysed datasets fail in one or another of our methods and do not allow to reliably estimate AI detectors. We encourage researchers to propose their own ways for quality assessment, which will allow to create a comprehensive system of evaluation of the detection datasets.
Our work aims to contribute to a better understanding of the difference between human and machine text, which will ultimately contribute to preserving the integrity of information in the world.

\section{Limitations}

In our work we focused on the task of binary classification, thus suggested methods are not optimal for the task of detection of the hybrid AI-human content. Also, some methods do not work properly on short texts, however, this is a known issue for short texts.

\bibliography{aaai25}
\appendix

\section{Data Description}
\label{appendix:data}
More detailed description with information on sources, topics and years of the datasets selected in this paper from competitions and research papers in Table~\ref{table:extend_info_datasets}

\begin{table*}[ht]
\centering
\begin{tabular}{@{}p{1.7cm}p{1.0cm}p{6.0cm}p{6.0cm}@{}}
\toprule
\textbf{Dataset} & \textbf{Year} & \textbf{Themes} & \textbf{Sources} \\ 
\midrule
\multicolumn{4}{c}{\textbf{Research papers datasets}}\\
\midrule
GPT2 & 2019 & WebText & GPT-2  \\
\hline
TweepFake & 2019 & Tweets  &  Markov Chains, RNN, LSTM, GPT-2   \\
\hline
HC3 & 2023 & ELI5, WikiQA, Wikipedia, Medicine, Finance & ChatGPT \\
\hline
GhostBuster & 2023 & Student Essays, News Articles, Creative Writing & ChatGPT, Claude \\
\hline
MGTBench & 2024 & Student Essays, News Articles, Creative Writing & ChatGPT. ChatGLM, Dolly, GPT4All, StableLM, Claude  \\
\hline
MAGE & 2024 & Opinions, Reviews, News, QA, Story Generation, Commonsense Reasoning, Knowledge Illustration, Scientific Writing & text-davinci-002, GPT-3.5, ChatGPT, LLaMA, GLM-130B, FLAN-T5, OPT, BLOOM, GPT-J-6B,
GPT-NeoX-20B  \\
\hline
M4 & 2024 & Wikipedia, Reddit ELI5, WikiHow, PeerRead, arXiv abstract & GPT-3.5, ChatGPT, Cohere, Dolly-v2, BLOOM  \\
\hline
OutFox & 2024 & Student Essays  & ChatGPT, GPT-3.5, FLAN-T5   \\
\midrule
\multicolumn{4}{c}{\textbf{Shared tasks datasets}}\\
\midrule
RuATD & 2022 & News, Social media, Wikipedia, Strategic Documents, Diaries & M-BART, M2M-100, OPUS-MT, mT5, ruGPT2, ruGPT3, ruT5-Base \\
\hline
DAGPap & 2022 & Scopus papers & Longformer Encoder-Decoder, GPT-3, Spinbot, GPT-Neo \\
\hline
AuTex & 2023 & Legal documents, Social media, How-to articles &  BLOOM, GPT-3, GPT-3.5 \\
\hline
IberAuTex & 2024 & News, Reviews, Emails, Essays, Dialogues, Wikipedia, Wikihow, Tweets & GPT-2, LLaMA, Mistral, Cohere, Claude, MPT, Falcon  \\
\hline
PAN & 2024 & News & Alpaca, BLOOM, Gemini, ChatGPT, gpt-4-turbo, LLaMA-2, Mistral, Qwen1.5, GPT-2  \\
\hline
SemEval Mono & 2024 & Wikipedia, WikiHow,
Reddit, arXiv, PeerRead, Student Essays &  GPT-3.5, GPT-4, Cohere, Dolly-v2, BLOOMz \\
\hline
SemEval Multi & 2024 & Wikipedia, WikiHow,
Reddit, arXiv, PeerRead, Student Essays, News & ChatGPT, GPT-3.5, GPT-4, LLaMA2, Cohere, Dolly-v2, BLOOM, Jais  \\
\hline
MGT-1 Mono & 2025 & CNN, DialogSum, Wikipedia, WikiHow, Eli5, Finance, XSum, PubMed, SQuAD, IMDb, 
Reddit, arXiv, PeerRead &  text-davinci-002, GPT-3.5, ChatGPT, OPT, LLaMA3, BLOOM, FLAN-T5, Cohere, Dolly, Gemma, Mixtral  \\
\hline
MGT-1  Multi & 2025 & CNN, DialogSum, Baike, WikiQA, WikiHow, Eli5, Finance, Psychology, XSum, PubMed, SQuAD, IMDb, 
Reddit, arXiv, PeerRead & text-davinci-002, GPT-3.5, ChatGPT, gpt4o, GLM, GPT-J, GPT-Neo, OPT, LLaMA2, LLaMA3, BLOOM, FLAN-T5, Cohere, Dolly, Gemma, Mixtral, Jais  \\
\bottomrule
\end{tabular}
\caption{More detailed descriptive statistics about domains and generators of the chosen datasets from competitions and research papers. ChatGPT is gpt-3.5-turbo, GPT-3.5 is text-davinci-003.}
\label{table:extend_info_datasets}
\end{table*}

\section{Evaluation results of competitions}
\label{appendix:evaluation}

Table \ref{table:competition_results} shows the winning scores in the competitions reviewed in this paper. In the AuTex and IberAuTex competitions it was forbidden to use additional data to fine-tune the detection algorithms. In the other collections it was allowed, we can notice a high quality near perfect in them. We should note the low value of the metrics on the RuATD dataset, which can be explained by the limited number of high-quality language models available in Russian during the competition.

\begin{table}[H]
\centering
\begin{tabular}{lcc}
\toprule
\textbf{Competition} & \textbf{Metric} & \textbf{Best result} \\ 
\midrule
RuATD  & Accuracy  & 0.820  \\ \hline
AuTex23-en  & Macro-F1  & 0.809  \\ \hline
AuTex23-es  & Macro-F1 &   0.708  \\ \hline
IberAuTex  & Macro-F1  & 0.805                                 \\ \hline
\makecell[l]{SemEval24 \\Mono}                      & Accuracy                                     & 0.975                                \\ \hline
\makecell[l]{SemEval24 \\Multi}  & Accuracy                                     & 0.959                                \\ \hline
PAN24                               & Avg. of 5 metrics*                                           & 0.924                                 \\ \hline
DAGPap22                            & Avg. F1-score                                & 0.994                                 \\
\hline
MGT-1 Mono & Macro-F1 & 0.8307 \\
\hline
MGT-1 Multi & Macro-F1   & 0.7916 \\
\bottomrule
\end{tabular}
\caption{Best results from each analysed competition. PAN24 used mean of 5 metrics, such as accuracy, F1 and other to evaluate \textit{efficiency} of the system.}
\label{table:competition_results}
\end{table}

\begin{figure*}[ht]
\centering
\includegraphics[width=\textwidth]{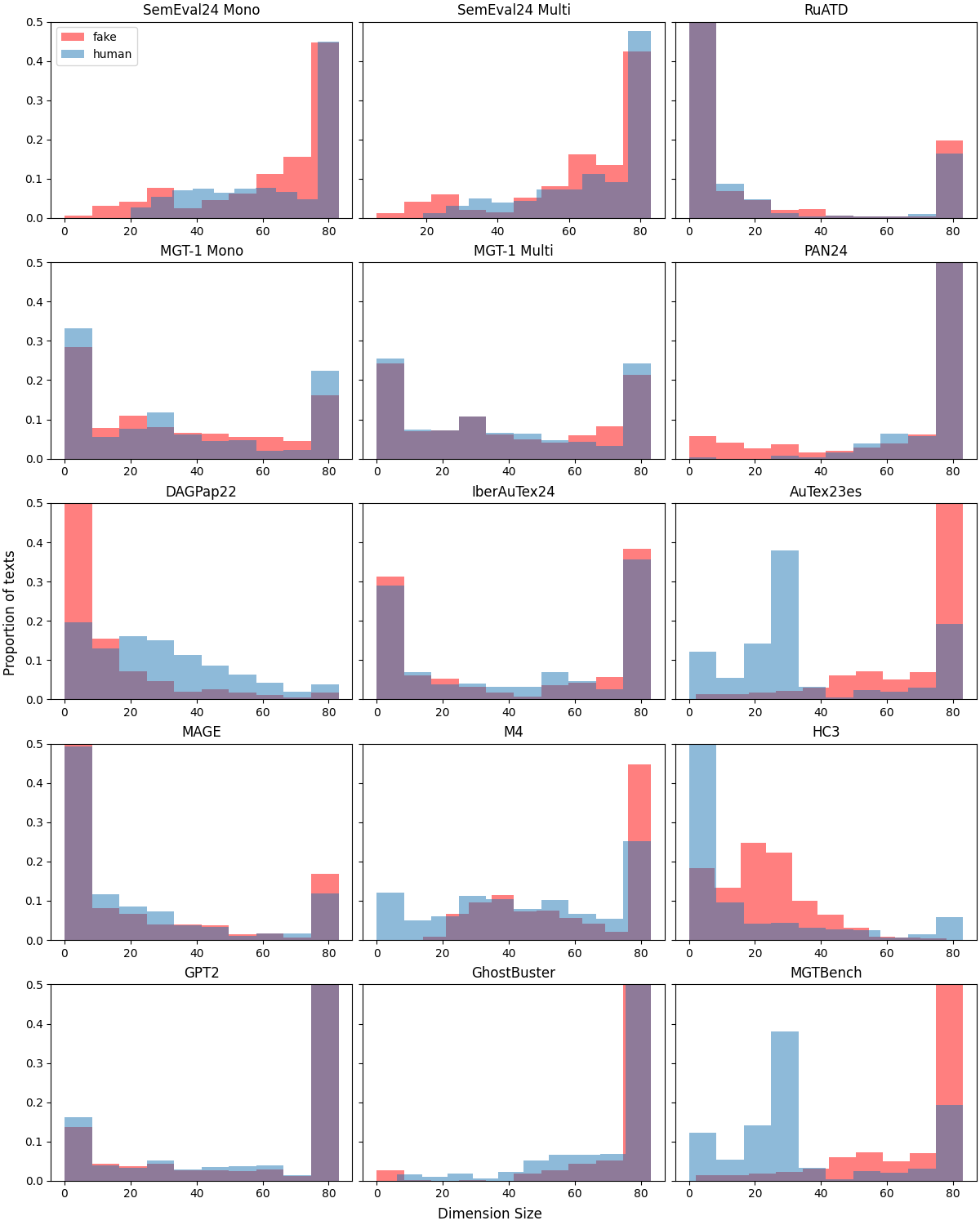}
\caption{Topological Time Series on all datasets.}
\label{fig:tts}
\end{figure*}

\section{Hyperparameters}
\label{appendix:hyperparameters}

\begin{table}[H]
\centering
\begin{tabular}{lc}
\toprule
\textbf{Hyperparameters} & \textbf{Values}\\
\hline
Epochs & 5* \\
Learning rate (LR) & 5e-5 \\
Warm-up steps & 50 \\
Weight decay & 0.01 \\
\bottomrule
\end{tabular}
\caption{Hyperparameters for fine-tuning mDeBERTa-base. We trained for 5 epochs with possibility of early exit.}
\label{tab:hyperparameters}
\end{table}
The training was carried out on NVIDIA GeForce RTX 3090. See hyperparameters in Table~\ref{tab:hyperparameters}.

\end{document}